\documentclass[letterpaper,10pt,journal,twoside]{IEEEtran}
\IEEEoverridecommandlockouts

\usepackage{iros}
\usepackage{bm}
\usepackage{amsmath}

\acrodef{dataset}[WBM3D]{Whole-Body Motions in 3D Scenes}
\acrodef{vkc}[VKC]{Virtual Kinematic Chain}
\acrodef{urdf}[URDF]{Unified Robot Description Format}
\acrodef{usd}[USD]{Universal Scene Description}
\acrodef{sdf}[SDF]{Signed Distance Function}

\let\oldnl\nl
\newcommand{\nosemic}{\SetEndCharOfAlgoLine{\relax}}
\newcommand{\nonl}{\renewcommand{\nl}{\let\nl\oldnl}}
\newcolumntype{x}{>{\columncolor{MistyRose}}c}
\newcolumntype{y}{>{\columncolor{LightCyan1}}c}

\newcommand\blfootnote[1]{%
  \begingroup
  \renewcommand\thefootnote{}\footnote{#1}%
  \addtocounter{footnote}{-1}%
  \endgroup
}

\newcommand*{\supp}{\ding{51}}
\newcommand*{\notsupp}{\ding{55}}
\newcommand*{\benchmark}{M${}^{3}$Bench}
\newcommand*{\datatool}{M${}^{3}$BenchMaker}
\newcommand*{\mpnet}{M$\pi$Net}
\newcommand*{\mptf}{M$\pi$Former}
\newcommand*{\modmp}{ModMP[O]}
\newcommand*{\modmps}{ModMP[S]}
\newcommand*{\mdiffusion}{MDiffusion}

\title{\LARGE \bf \benchmark: Benchmarking Whole-body \underline{M}otion Generation for \\ \underline{M}obile \underline{M}anipulation in 3D Scenes}%

\author{Zeyu Zhang, Sixu Yan, Muzhi Han, Zaijin Wang, Xinggang Wang, Song-Chun Zhu, Hangxin Liu
}

\begin{document}

\twocolumn[{
\renewcommand\twocolumn[1][]{#1}
\maketitle
\thispagestyle{empty}
\pagestyle{empty}
\vspace{-0.5cm}
\begin{center}
    \centering
    \captionsetup{type=figure}
        \includegraphics[width=\linewidth]{figures/benchmark/benchmark.pdf}
        \caption{\textbf{The \benchmark~benchmark} challenges mobile manipulators to generate whole-body motion trajectories for object manipulation in 3D scenes. Given a 3D scan, a target segmentation mask, and a task description, the robot must understand its embodiment, environment, and task objectives to produce coordinated motions for picking or placing objects. 
        }  
    \label{fig:benchmark}
\end{center}
}]


\blfootnote{Manuscript received October 13, 2024; revised February 14, 2025; accepted May 9, 2025. Date of publication May xx, 2025; date of current version May xx, 2025. This paper was recommended for publication by Editor Aniket Bera upon evaluation of the Associate Editor and Reviewers' comments. This work was supported in part by the National Natural Science Foundation of China (No. 62376031). (\textit{Zeyu Zhang and Sixu Yan contributed equally to this work.})(\textit{Corresponding authors: Hangxin Liu.})}
\blfootnote{Zeyu Zhang, Zaijin Wang, Song-Chun Zhu, and Hangxin Liu are with the State Key Laboratory of General Artificial Intelligence, Beijing Institute for General Artificial Intelligence (BIGAI), Beijing 100080, China. (e-mails: zhangzeyu@bigai.ai; liuhx@bigai.ai;).}
\blfootnote{Sixu Yan and Xinggang Wang are with School of Electronic Information and Communications, Huazhong University of Science and Technology, Wuhan 430079, China (e-mails: yansixu@hust.edu.cn; xgwang@hust.edu.cn).}
\blfootnote{Muzhi Han is with the Center for Vision, Cognition, Learning, and Autonomy (VCLA), Statistics Department, University of California, Los Angeles (UCLA), Los Angeles, CA 90095, USA.}
\blfootnote{Song-Chun Zhu is also with the School of Artificial Intelligence and the Institute for Artificial Intelligence, Peking University, Beijing 100871, China.}

\begin{abstract}
We propose \benchmark, a new benchmark for whole-body motion generation in mobile manipulation tasks. Given a 3D scene context, \benchmark~requires an embodied agent to reason about its configuration, environmental constraints, and task objectives to generate coordinated whole-body motion trajectories for object rearrangement. \benchmark~features 30,000 object rearrangement tasks across 119 diverse scenes, providing expert demonstrations generated by our newly developed \datatool, an automatic data generation tool that produces whole-body motion trajectories from high-level task instructions using only basic scene and robot information. Our benchmark includes various task splits to evaluate generalization across different dimensions and leverages realistic physics simulation for trajectory assessment. Extensive evaluation analysis reveals that state-of-the-art models struggle with coordinating base-arm motion while adhering to environmental and task-specific constraints, underscoring the need for new models to bridge this gap. By releasing \benchmark~and \datatool~at \url{https://zeyuzhang.com/papers/m3bench}, we aim to advance robotics research toward more adaptive and capable mobile manipulation in diverse, real-world environments.
\end{abstract}
\begin{IEEEkeywords}
Mobile Manipulation, Whole-body Motion Generation, Embodied AI
\end{IEEEkeywords}

\section{Introduction}
\begin{figure}[t!]
    \centering
    \begin{subfigure}[b]{\linewidth}
        \centering
        \includegraphics[width=\linewidth]{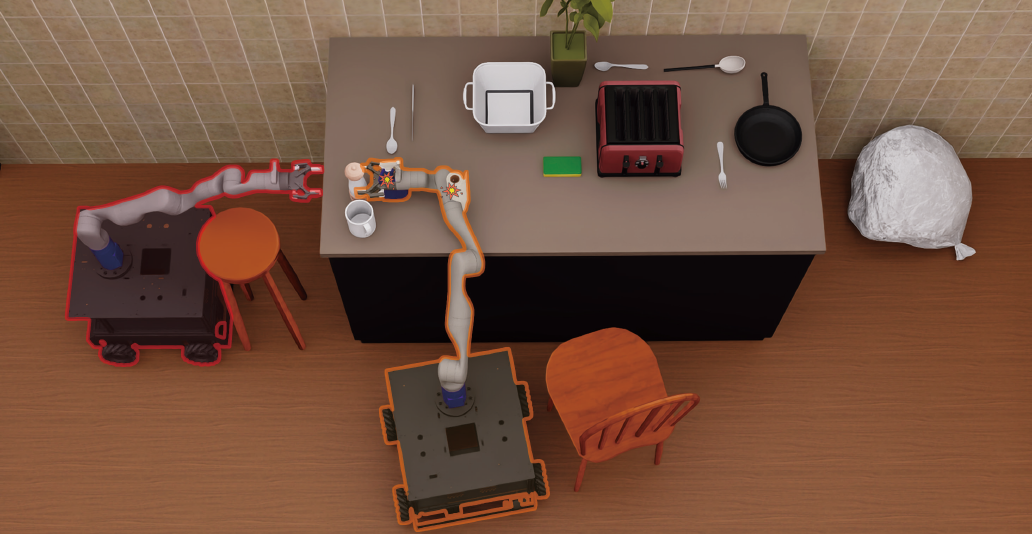}
        \caption{Failure scenarios of mobile manipulation in 3D scenes}
        \label{fig:teaser_fail_traj}
    \end{subfigure}
    \\
    \begin{subfigure}[b]{\linewidth}
        \centering
        \includegraphics[width=\linewidth]{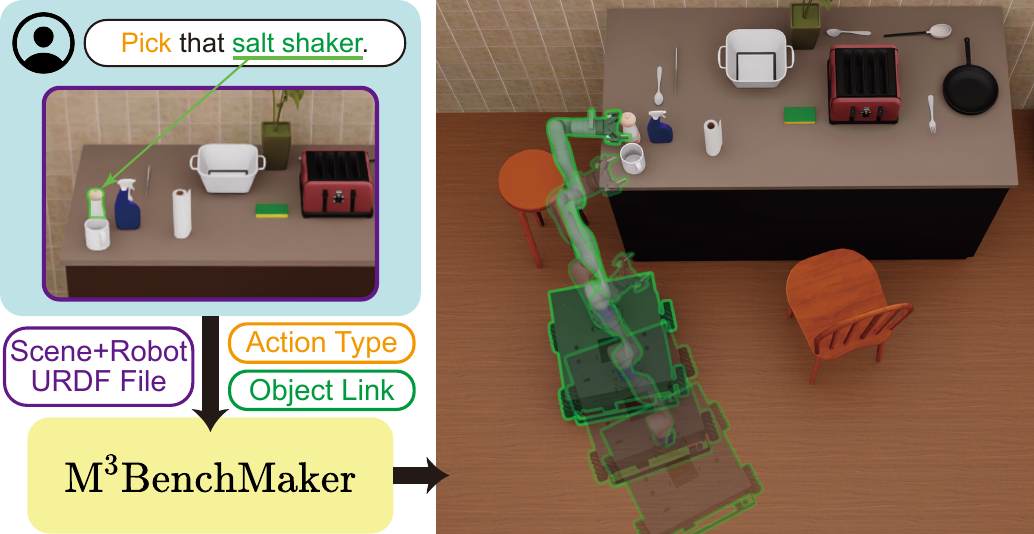}
        \caption{An overview of whole-body motion data collection tool}
        \label{fig:teaser_data_tool}
    \end{subfigure}
    \caption{\textbf{Illustration of whole-body motion trajectories in 3D scenes.} (a) Treating the mobile base and arm as separate entities can lead to two typical failures: a nearby navigable position may be impractical for the arm to reach the object (red), and a feasible grasp pose may be unachievable due to the robot's embodiment and environmental constraints (orange). (b) Our tool generates feasible whole-body motion trajectories from high-level instructions, requiring only the action type, target object, and URDF files of the scene and robot. The green overlay illustrates a generated trajectory for the ``pick that salt shaker'' task.}
    \label{fig:teaser}
\end{figure}

\IEEEPARstart{H}{umans} possess an innate ability to manipulate their environment with remarkable flexibility and coordination, seamlessly integrating locomotion and manipulation. In contrast, robots still struggle to achieve this level of adaptability and proficiency in mobile manipulation. Current learning-based models and motion planning methods for mobile manipulators often address individual subproblems in isolation, such as navigating to waypoints, manipulating with a fixed mobile base, or grasping objects. However, neglecting the potential of coordinated whole-body motion can lead to misalignment between module outputs and task constraints. For instance, in a typical object-fetching task, a navigable position near the target object may still be impossible for the arm to reach the object, or a feasible grasp pose may become unachievable due to collisions with surrounding objects (see~\cref{fig:teaser_fail_traj}). These limitations underscore the necessity of coordinating whole-body motion with a comprehensive understanding of robot embodiment, environmental context, and task objectives to enable effective mobile manipulation in complex 3D scenes.

To generate whole-body motion for mobile manipulation tasks, there is an ongoing debate regarding the effectiveness and limitations of model-based motion planning methods versus data-driven learning-based models. While motion planning can produce complex whole-body mobile manipulation skills~\cite{jiao2021consolidated}, its effectiveness and generalizability in real-world scenarios are constrained by its reliance on perfect environmental knowledge~\cite{mo2021o2oafford,han2022scene,han2021reconstructing,zhang2023part} or predefined goal configurations (\eg, grasp poses~\cite{gu2022multi,Li2022GenDexGrasp, zhang2022understanding}). On the other hand, learning-based models have shown promising results in execution under perceptual and action noise, chaining primitive skills, and adapting to certain environmental variations~\cite{xie2023part}. However, they have yet to demonstrate robust base-arm coordination with situated goal configurations (\eg, achieving specific grasps or object placements). Learning complex mobile manipulation tasks requires datasets that capture whole-body motions in 3D scenes, yet such datasets remain scarce due to the challenges in generating whole-body motion data. Furthermore, evaluating learned models necessitates a standardized environment for fair benchmarking.

Table~\ref{tab:benchmark_comparison} presents recent state-of-the-art benchmarks for Embodied AI and robotics. Many of these benchmarks simplify actions to symbolic operations~\cite{wang2024llm, song2023llmplanner, jiao2021efficient} or navigation~\cite{anderson2018vision}, lacking physical interaction with the environment. While more recent benchmarks enable fixed-base manipulators to interact with objects in realistic simulations~\cite{james2020rlbench,mu2021maniskill,chamzas2021motionbenchmaker,gu2022maniskill2} or allow mobile agents to navigate and manipulate in 3D scenes~\cite{szot2021habitat, zhi2024closed, yan2025m, liu2024pr2}, they often overlook the necessity of coordinating base and arm motions.

To address the need for generating whole-body motions in mobile manipulation, we introduce \benchmark, a comprehensive benchmark that features challenging object rearrangement tasks that require a mobile manipulator to reason about its embodiment, environmental context, and task objectives to generate coordinated motions for picking and placing objects in diverse household scenes (see~\cref{fig:benchmark}). \benchmark~comprises 30,000 object rearrangement tasks involving 32 distinct object types across 119 household scenes, covering a broad spectrum of task objectives and environmental constraints relevant to embodied mobile manipulation. Additionally, it includes rich metadata, such as natural language task instructions, panoptic maps, and egocentric camera videos, making it a valuable resource for related research in Embodied AI, such as embodied instruction following and human-AI collaboration.

Leveraging \benchmark, we developed \datatool~(see \cref{fig:teaser_data_tool}), an automatic data generation tool designed to produce whole-body motion trajectories as expert demonstrations for robot learning. \datatool~procedurally generates coordinated trajectories from high-level task instructions, requiring only the action type, object link, and the \ac{urdf} of the scene and robot. It employs an energy-based model to predict grasp pose and placement candidates~\cite{urain2023se}, and it leverages an advanced virtual kinematics technique~\cite{jiao2021consolidated} to compute coordinated whole-body motion trajectories (see~\cref{sec:datatool} for details). This tool not only addresses the scarcity of high-quality whole-body mobile manipulation data but also allows researchers to generate additional samples customized to specific robot and scene configurations for their own studies.

To enable in-depth evaluation of motion generation from 3D scans for mobile manipulation, \benchmark~incorporates various task splits to assess generalization across different dimensions, such as novel scenes and objects. We utilize a realistic physics simulation platform~\cite{makoviychuk2isaac} to evaluate the feasibility of generated motion trajectories, ensuring that the robot can physically grasp objects and place them stably at the desired locations. Furthermore, our benchmarking reveals that sampling- and optimization-based motion planning methods~\cite{schulman2013finding, schulman2014motion}, even when augmented with affordance prediction, as well as learning-based autoregressive planning and generative AI techniques~\cite{xie2023part, fishman2023motion, zhang2023learning}, struggle to effectively solve mobile manipulation tasks when required to account for goals such as grasp poses for picking and placement locations for placing actions. After integrating action goals into motion generation, learning-based methods outperform modularized motion planning in computational efficiency and simplicity of problem setup but still lag in motion accuracy. This underscores the importance of high-quality whole-body mobile manipulation data generated by tools like \datatool~and highlights the necessity of \benchmark~for advancing research in whole-body motion generation for mobile manipulation in 3D scenes.

\begin{table*}[th!]
    \centering
    \footnotesize
    \setlength{\tabcolsep}{7pt}
    \caption{\textbf{Relevant datasets and benchmarks in robotics.} The \benchmark~provides comprehensive whole-body motion demonstrations for object manipulation across 566 household scenes. \textbf{Mobile Manipulation}: Simultaneous navigation and object manipulation with foot-arm coordination. \textbf{Whole-body Demonstration}: Provides whole-body motion data. ${}^1$Simplified cases without navigation and coordination. \textbf{Procedural Generation}: Algorithmic procedure for creating varied tasks and trajectories. \textbf{Household Scene}: Tasks performed in 3D household environments. \textbf{Language}: Natural language task descriptions. \textbf{Physical Grasp}: Realistic physics-based grasping simulation. ${}^2$Simplified grasp (\eg, attach). \textbf{Egocentric Perception}: Provides egocentric visual sensory data (\eg, RGB-D images). ${}^3$No rendered RGB images. \textbf{Flexible Material}: Customizable materials and textures for visual diversity.}
    \begin{tabular}{lccccccccc}
    \toprule
    Benchmark & \thead{Mobile\\ Manipulation} & \thead{Whole-body\\ Demonstration} & \thead{Procedural\\ Generation} & \thead{Household\\ Scene} & \thead{Language} & \thead{Physical\\ Grasp} & \thead{Egocentric\\ Perception} & \thead{Flexible\\ Material} \\
    \midrule
    ACRV~\cite{leitner2017acrv} & \notsupp & \notsupp & \notsupp & \notsupp & \notsupp & \supp & \supp & \notsupp \\
    Alfred~\cite{ALFRED20} & \notsupp & \notsupp & \supp & 120 & \supp & \notsupp & \supp & \notsupp \\
    ManiSkill~\cite{mu2021maniskill, gu2022maniskill2} & \supp & $\text{\supp}^1$ & \notsupp & \notsupp & \notsupp & \supp & \supp & \notsupp \\
    Calvin~\cite{mees2022calvin} & \notsupp & \notsupp & \supp & \notsupp & \supp & \supp & \supp & \notsupp \\
    Behavior~\cite{srivastava2022behavior} & \supp & \notsupp & \supp & 50 & \notsupp & \notsupp & \supp & \supp \\
    RLBench~\cite{james2020rlbench}  & \notsupp & \notsupp & \supp & \notsupp & \notsupp & $\text{\supp}^2$ & \supp & \notsupp \\
    VLMbench~\cite{zheng2022vlmbench} & \notsupp & \notsupp & \supp & \notsupp & \supp & $\text{\supp}^2$ & \supp & \notsupp \\
    Ravens~\cite{zeng2021transporter} & \notsupp & \notsupp & \supp & \notsupp & \supp & $\text{\supp}^2$ & \supp & \notsupp \\
    MotionBenchMaker~\cite{chamzas2021motionbenchmaker} & \notsupp & \notsupp & \supp & \notsupp & \notsupp & \notsupp & $\text{\supp}^3$ & \notsupp \\
    Habitat HAB~\cite{szot2021habitat} & \supp & \notsupp & \supp & 105 & \notsupp & \notsupp & \supp & \notsupp \\
    ARNOLD~\cite{gong2023arnold} & \notsupp & \notsupp & \notsupp & 20 & \supp & \supp & \supp & \supp \\
    \midrule
    Ours & \supp & \supp & \supp & 119 & \supp & \supp & \supp & \supp \\
    \bottomrule
    \end{tabular}
    \label{tab:benchmark_comparison}
\end{table*}

\paragraph*{Contribution} We make the following contributions: 
\begin{itemize}
    \item We introduce \benchmark~for benchmarking task-oriented whole-body motion generation for mobile manipulation in household environment, and we provide assets required for testing traditional planning-based methods or learning-based methods. 
    \item We develop \datatool, an automatic whole-body motion generation tool based on high-level task instructions, which can be easily customized for different robot and scene configurations.
    \item We provide an in-depth evaluation of motion generation from 3D scans for mobile manipulation, revealing weaknesses of current arts in promoting future research in mobile manipulation across diverse 3D scenes.
\end{itemize}

\paragraph*{Overview} The remainder of this paper is organized as follows. \cref{sec:datatool} describes the key components of \datatool. \cref{sec:benchmark} details the development of the environment and the benchmarking setup. We implement multiple methods for mobile manipulation and discuss their performance in \cref{sec:exp}, and we conclude the paper in \cref{sec:conclude}.

\section{The \datatool} \label{sec:datatool}

\begin{figure*}[t!]
    \centering
    \includegraphics[width=\linewidth]{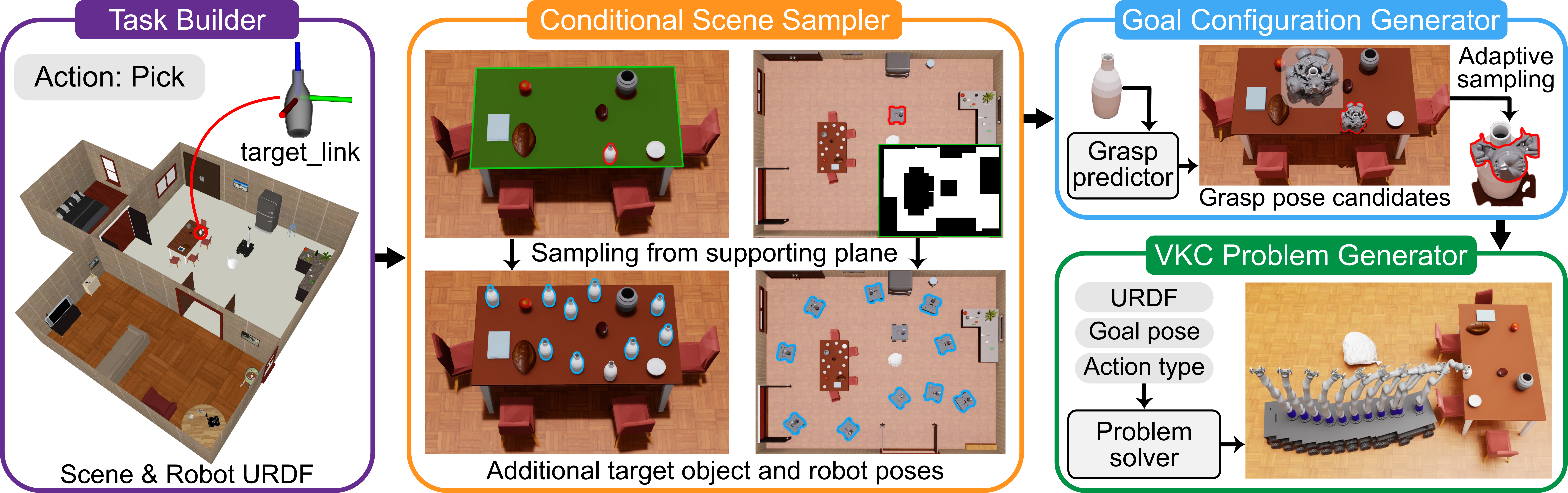}
    \caption{\textbf{Overview of the \datatool.} The \textbf{Task Builder} allows users to specify manipulation tasks via high-level definitions using \ac{urdf}, target object link, and action. The \textbf{Conditional Scene Sampler} augments data by generating object and robot poses (blue outline) in terms of their supporting planes (green outline) of target objects (red outline). The \textbf{Goal Configuration Generator} produces task-specific goal poses using a pre-trained model for grasp/placement candidates. The \textbf{VKC Problem Generator} constructs optimization programs for computing whole-body motion trajectories that satisfy task objectives and constraints via \ac{vkc}~\cite{jiao2021consolidated}.}
    \label{fig:data_tool}
\end{figure*}

Diverse whole-body motion trajectories for mobile manipulators in complex 3D environments is crucial for advancing embodied AI. 
However, collecting expert demonstrations for training models are usually time-consuming and challenging. 
To address this, we introduce \datatool, a user-friendly tool that streamlines the generation of whole-body motion trajectories in 3D scenes, significantly reducing the time and effort required to create large-scale datasets for mobile manipulation tasks in various environments.
Notably, \datatool~is adaptable to different robot and scene given the \ac{urdf} files that describe the configurations, enabling researchers to generate customized whole-body motion trajectories for their specific research needs. 
\cref{fig:data_tool} illustrates the architecture of the \datatool.

\subsection{Task Builder}

The task builder serves as the primary user interface, allowing users to define manipulation tasks using high-level action commands such as picking, placing, and reaching. 
Users no longer need to manually specify grasping poses, placement locations, base positions, or create optimization programs for motion trajectories.
To define a task, users simply select target object links from the scene \ac{urdf}, set the robot's initial position, and specify the desired action types.
The task builder then creates an instance of the data generation pipeline, integrating subsequent modules to procedually generate whole-body motion trajectories. 
For enhanced data diversity, the task builder supports data augmentation via the Conditional Scene Sampler (see \cref{subsec:scene_sampler}). 
This feature facilitates the training and evaluation of embodied AI models in complex environments by generating varied scenarios from a single task definition.

\subsection{Conditional Scene Sampler}\label{subsec:scene_sampler}

The conditional scene sampler generates diverse initial configurations for data augmentation by randomizing object and robot positions and orientations. 
It produces variations dependent on the original scene's object relations, ensuring physical feasibility and contextual consistency required by the task.
For instance, in a task involving picking an object from a table, the sampler ensures the sampled objects remains on table (see~\cref{fig:data_tool} orange box). 
This is achieved by recognizing supporting planes for objects and the robot through analysis of surrounding geometries.

To identify supporting planes, we parameterize a surface plane as $\boldsymbol{\pi} = \left\langle\boldsymbol{n}^T, d, U \right\rangle$, where $\boldsymbol{n} \in \mathbb{R}^3$ is the normal vector, $d$ is the distance to origin, and $U = \{\boldsymbol{u}| \boldsymbol{u} \in \mathbb{R}^3\}$ defines the plane's polygon outline. The most likely supporting plane $\boldsymbol{\pi}_s$ for a bottom surface $\boldsymbol{\pi}_o$ is identified by solving:
\begin{align}
    \operatorname*{argmax}_{\boldsymbol{\pi}_s \in \prod}&~\text{A}\left(U_s \cap \text{proj}_{o, s}(U_o)\right) / \text{A}(U_o), \label{eqn:supp_plan_obj} \\
    \text{s.t.}~&\frac{1}{\left|U_{o}\right|} \sum_{\boldsymbol{u} \in U_{o}} \boldsymbol{n}_{s}^{T} \boldsymbol{u}+d_{s} \leqslant \theta_d, \label{eqn:supp_plan_align} \\
    & \operatorname{abs}\left(\boldsymbol{n}_{p}^{i T} \boldsymbol{n}_{c}^{j}\right) \geqslant \theta_{a}. \label{eqn:supp_plan_dist}
\end{align}
where $\prod$ is a set of supporting plane candidates, $\text{A}(\cdot)$ denotes polygon area, $\cap$ computes intersection, and $\text{proj}_{o, s}(U_o) = \{\boldsymbol{u} - \left(\boldsymbol{n}_{s}^{T} \boldsymbol{u}+d_{s}\right) \boldsymbol{n}_{s} | \boldsymbol{u} \in U_o\}$ projects bottom surface points onto the supporting plane, $\theta_d$ and $\theta_a$ are distance and angle thresholds. \cref{eqn:supp_plan_obj} defines the contact ratio, while \cref{eqn:supp_plan_align,eqn:supp_plan_dist} enforce alignment and distance constraints.
The complete sampling procedure is detailed in~\cref{alg:conditional_scene_sampler}.
We utilize the method in \cite{han2022scene} to extract surface planes and solve the optimization problem by iteratively identifying the plane that maximizes \cref{eqn:supp_plan_obj} while satisfying the constraints. 

\begin{algorithm}[tb!]
    \caption{Conditional Scene Sampler}
    \label{alg:conditional_scene_sampler}
    \LinesNumbered
    \SetKwInOut{KIN}{Input}
    \SetKwInOut{KOUT}{Output}
    
    \KIN{Target object $\boldsymbol{\pi}_o$, candidate planes $\prod$, thresholds $\theta_d, \theta_a$, number of samples $N$}
    \KOUT{A set of feasible object poses $\mathcal{S}$}
    
    $\mathcal{S} \leftarrow \emptyset$ \;

    \nonl\nosemic \textcolor{blue}{// Filter candidates in terms of \cref{eqn:supp_plan_align,eqn:supp_plan_dist}} \;
    $\prod_c \leftarrow \texttt{FilterSupportPlane}(\boldsymbol{\pi}_o, \prod, \theta_d, \theta_a)$ \;
    \nonl\nosemic \textcolor{blue}{// Determine supporting plane according to \cref{eqn:supp_plan_obj}} \;
    $\boldsymbol{\pi}_s \leftarrow \texttt{CalcSupportPlane}(\boldsymbol{\pi}_o, \prod_c)$ \; 

    \While{$\mathcal{S}.\texttt{size}() < N$}{
        $p \leftarrow \texttt{samplePoseOnPolygon}(\boldsymbol{\pi}_s)$ \;

        \nonl\nosemic \textcolor{blue}{// Check if sampled pose is within plane $\boldsymbol{\pi}_s$} \;
        \If{$\texttt{withinPolygon}(\boldsymbol{\pi}_o, p, \boldsymbol{\pi}_s)$}{
            $\mathcal{S}.\texttt{add}(p)$ \textcolor{blue}{// add sampled pose to $\mathcal{S}$} \;
        }
    }
    
    \Return $\mathcal{S}$ \;
\end{algorithm}

\subsection{Goal Configuration Generator}

This module efficiently generates 6D end-effector poses for grasping or placing target objects, serving as optimization objectives for motion planning. 
We employ an energy-based model to predict candidate goal configurations based on target object geometry~\cite{urain2023se}. 
However, this object-centric approach, which considers only object geometry without accounting for the robot's kinematic constraints or environmental contexts, results in only a small subset of candidates being feasible for the task.
To address the computational expense of evaluating all candidates through motion planning, we developed an adaptive sampling algorithm that efficiently draws samples from the candidate set, significantly accelerating the motion generation process.

Detailed in \cref{alg:adaptive_sampling}, our algorithm iteratively selects and updates the sampling probability of candidates based on their feasibility scores. 
It utilizes a K-D tree for efficient neighbor search and initializes feasibility scores using the candidates' energy values. 
When a candidate fails the feasibility check, the feasibility scor es of its neighbors, identified via the K-D tree within a specified distance, are halved during the update.
By concentrating sampling in promising regions of the goal configuration space while maintaining exploration, the algorithm significantly reduces the number of expensive feasibility checks required to identify viable goal configurations.

\begin{algorithm}[tb!]
    \caption{Adaptive Goal Sampling}
    \label{alg:adaptive_sampling}
    \LinesNumbered
    \SetKwInOut{KIN}{Input}
    \SetKwInOut{KOUT}{Output}
    
    \KIN{candidate set $\mathcal{C}$}
    \KOUT{goal configuration $g$}
    
    \nosemic $T \leftarrow \texttt{KDTree}(\mathcal{C})$ \; \label{alg:sampling:line:kdtree}
    \nosemic $scores \leftarrow \texttt{initFeasibilityScore}(\mathcal{C})$\; \label{alg:sampling:line:score_init}
    
    \While{feasible goal not found}{
        \nonl\nosemic \textcolor{blue}{// calculate probability for each candidate} \;
        \nosemic $probs \leftarrow \texttt{calcSamplingProb}(\mathcal{C}, scores)$\;
        \nonl\nosemic \textcolor{blue}{// draw a single condidate index from distribution} \;
        \nosemic $i \leftarrow \texttt{drawSample}(\mathcal{C}, probs)$ \; \label{alg:sampling:line:sampling}

        \nonl\nosemic \textcolor{blue}{// Check feasibility of sampled candidate}\;
        \If{$\texttt{checkFeasibility}(\mathcal{C}[i])$}{ \label{alg:sampling:line:feasibility}
            \nonl\nosemic \textcolor{blue}{// found feasible configuration} \;
            \nosemic $g \leftarrow \mathcal{C}[i]$\;
            \nosemic \textbf{break} \;
        }
        
        \nonl\nosemic \textcolor{blue}{// Update feasibility scores in neighbors}\;
        \nosemic $neighbors \leftarrow T.\texttt{GetNeighbors}(\mathcal{C}[i])$ \;
        \nosemic $scores[neighbors] \leftarrow scores[neighbors] \times 0.5$ \; \label{alg:sampling:line:update_score}
        
        \nonl \textcolor{blue}{// update K-D tree and remove checked candidate}\;
        \nosemic $T \leftarrow \texttt{UpdateKDTree}(T, \mathcal{C}[i])$\;
        \nosemic $\mathcal{C}.\texttt{remove}(i)$\;
    }
    \Return $g$ \;
\end{algorithm}

\subsection{VKC Problem Generator} \label{subsec:prob_gen}

The \ac{vkc} problem generator automates the construction of motion planning programs, formulating comprehensive optimization problems that encapsulate all necessary constraints and objectives for computing whole-body motion trajectories, utilizing task specifications, \ac{urdf}, and goal configurations from preceding modules.
We employ the \ac{vkc} approach~\cite{jiao2021consolidated} to solve for whole-body motion of mobile manipulators, modeling the mobile base, robot arm, and manipulated object as a unified system, achieving superior foot-arm coordination through simultaneous optimization and surpassing traditional methods that separate base and arm planning.

Our implementation follows TrajOpt and ROS-Industrial Tesseract conventions~\cite{armstrong_2018}, effectively incorporating kinematic constraints while avoiding large-space searches. 
The trajectory optimization minimizes joint travel distances and overall smoothness, with inequality constraints for joint limits, collision avoidance, and end-effector pose reaching. 
We adopt a sequential convex optimization method~\cite{schulman2014motion} to solve the resulting problem, yielding feasible, coordinated whole-body motion trajectories for diverse mobile manipulation tasks without manual task-specific planner programming.

By automating these processes, \datatool~empowers researchers to efficiently collect tailored whole-body motion trajectories, significantly advancing embodied AI in complex 3D environments.

\section{The \benchmark}\label{sec:benchmark}

The \benchmark~aims to advance robot capabilities in coordinating whole-body movements within complex environments, inspired by human ability to seamlessly perform such tasks.
It challenges mobile manipulators to generate coordinated whole-body motion trajectories for picking or placing everyday objects in 3D scenes, requiring agents to jointly understand their embodiment, environmental contexts, and task objectives from 3D scans.

\subsection{Simulation Environment}

\textbf{Simulation Platform.} 
Our benchmark, built on Isaac Sim~\cite{makoviychuk2isaac}, provides a high-fidelity physics simulation that meticulously models real-world properties and interactions. 
This platform enables precise evaluation of motion trajectory feasibility, grasping abilities, and the complex interplay between mobility and manipulation. 
Additionally, it could generate rich perceptual data (\eg, RGB-D image) that closely mimics the sensory input available to real-world robots.

\textbf{Scene and Robot Configuration.} 
The benchmark comprises 119 diverse household scenes containing 32 types of objects, curated from PhyScene~\cite{yang2024physcene}. 
These interactive 3D scenes are enhanced with physical properties and rich materials for photo-realistic and physics-realistic simulation. 
For the robot, we employ a common mobile manipulator configuration: a 7-DoF Kinova Gen3 robotic arm with a parallel gripper, mounted on an omnidirectional mobile base. This setup facilitates complex manipulations requiring coordinated base and arm movements.

\subsection{Task Design and Variations}

\benchmark~focuses on two primary object rearrangement tasks: picking and placing. 
Given a 3D point cloud of the scene, a mask of the target object, and its initial configuration, the robot must generate whole-body motion trajectories to manipulate the object. 
The tasks are defined as: (i) \textit{Pick tasks}: Navigate to, reach, and grasp a specified object from its initial location; (ii) \textit{Place tasks}: Transport a held object to a designated location and place it stably.
Success in both tasks requires avoiding collisions along the trajectory and maintaining the desired goal state for 2 seconds.

The task pool encompasses a wide range of mobile manipulation scenarios, featuring 32 object types with varying properties across 119 diverse household scenes. 
Each scene presents unique layouts, furniture arrangements, and obstacle configurations.
Tasks are generated by selecting appropriate objects and placement locations based on scene categories.
We employ the conditional scene sampler (\cref{subsec:scene_sampler}) to generate various initial configurations, further challenging the robot to generate coordinated whole-body motions while adapting to environmental constraints and task objectives.

\subsection{Data Collection}

\textbf{Demonstration Generation.} 
We utilized our developed \datatool~to generate demonstrations for each task. 
The tool takes as input the scene and robot \ac{urdf}, target object link, and task type (pick or place), then generates a whole-body motion trajectory for the robot.
The optimization program in \datatool~ensures these trajectories are collision-free and kinematically feasible.
Each trajectory is then verified for physical feasibility in Isaac Sim, with only valid demonstrations and their corresponding tasks included in the benchmark. 
In total, we collected 30k valid demonstrations, each containing 30 waypoints.

\begin{figure}[t!]
    \centering
    \begin{subfigure}[b]{\linewidth}
        \centering
        \includegraphics[width=\linewidth]{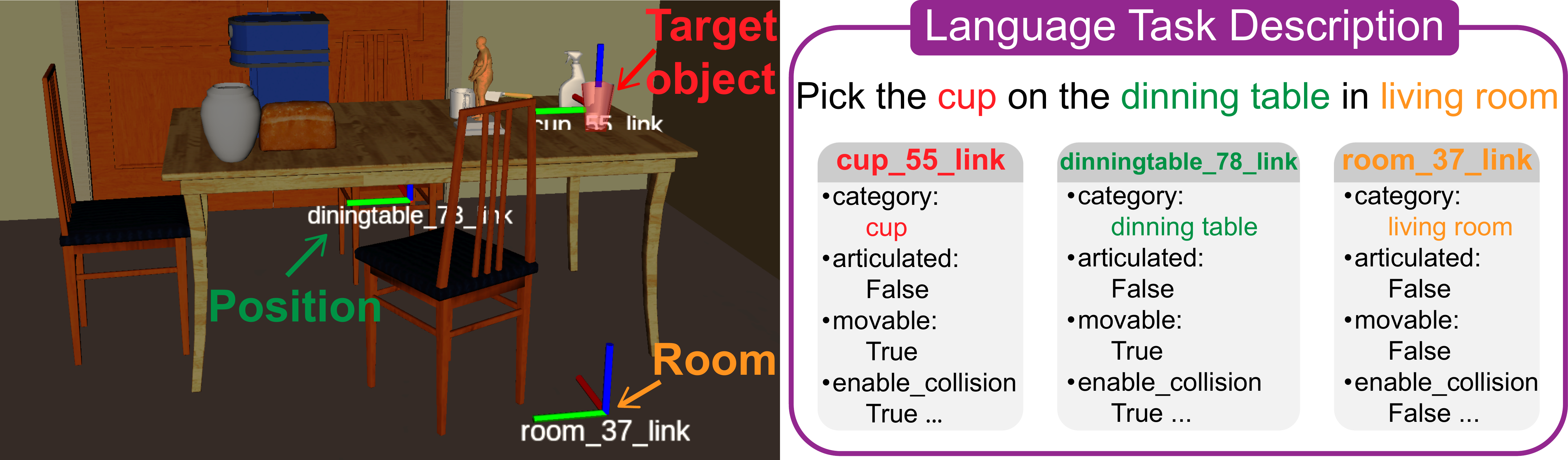}
        \caption{An example of \ac{urdf} annotation and language task description.}
        \label{fig:metadata_lang}
    \end{subfigure}
    \\
    \begin{subfigure}[b]{\linewidth}
        \centering
        \includegraphics[width=\linewidth]{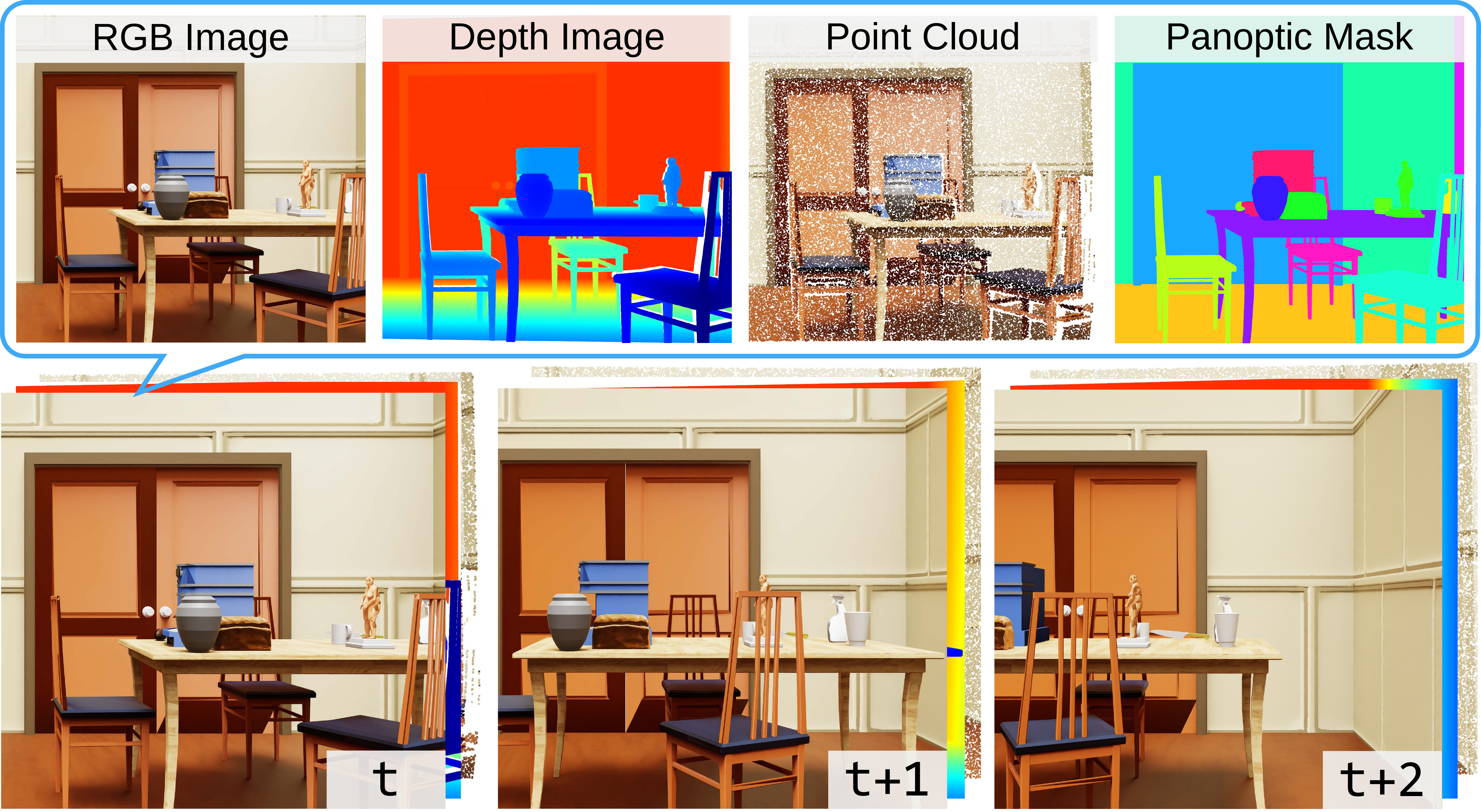}
        \caption{An example of sequential egocentric views of the robot.}
        \label{fig:metadata_ego}
    \end{subfigure}
    \caption{\textbf{An illustration of metadata.}}
    \label{fig:metadata}
\end{figure}

\textbf{Additional Metadata.} 
To facilitate embodied AI research, we provide comprehensive metadata for each task (see~\cref{fig:metadata}). 
This includes annotations for all links in the scene \ac{urdf}, covering object categories and simulation properties. 
We employ a template-based approach with lexicalized phrase candidates to generate language instructions for each task.
For example, the template ``Pick [object] in [room] on [position]'' might be realized as ``Pick the cup in the living room on the dining table''.
During task execution, Isaac Sim's built-in rendering capabilities, combined with annotated information, generate pixel-accurate semantic and instance segmentations along with egocentric camera views.
This rich combination of annotations, trajectory data, and language instructions creates a comprehensive resource for exploring various aspects of embodied intelligence.

\subsection{Benchmark} \label{subsec:benchmark}

\begin{table}[t]
    \centering
    \footnotesize
    \begin{minipage}{0.52\linewidth}
        \centering
        \setlength{\tabcolsep}{5.5pt}
        \caption{Number of pick/place task samples in each data split.}
        \label{tab:benchmark_stats_1}
        \begin{tabular}{lcc}
        \toprule
        Split & \textit{Pick} & \textit{Place} \\
        \midrule
        \textit{Train} & 14,793 & 7,478 \\
        \textit{Val} & 948 &  479 \\
        \textit{Test} & 3,225 & 1,630 \\
        \textit{Novel Object} & 688 & 397 \\
        \textit{Novel Scene} & 762 & 369 \\
        \textit{Novel Scenario} & 204 & 77 \\
        \midrule
        Total & 20,620 & 10,430 \\
        \bottomrule
        \end{tabular}
    \end{minipage}%
    \hfill
    \begin{minipage}{0.44\linewidth}
        \centering
        \setlength{\tabcolsep}{9pt}
        \caption{Number of rooms and target objects in \benchmark}
        \label{tab:benchmark_stats_2}
        \begin{tabular}{lc}
        \toprule
        Statistics & Value \\
        \midrule
        Bathroom & 132 \\
        Bedroom & 198 \\
        Kitchen & 97 \\
        Living room & 129 \\
        Total scenes & 119 \\
        \midrule
        Object types & 32 \\
        Total objects & 588 \\
        \bottomrule
        \end{tabular}
    \end{minipage}
\end{table}

\textbf{Data Split and Statistics.} 
The tasks in \benchmark~are carefully divided into several splits to assess different aspects of generalization capabilities.
Objects and scenes are randomly categorized into seen and unseen subsets.
The primary evaluation set, the \textit{Base} split, encompasses all seen objects and scenes, divided into \textit{Train} (75\%), \textit{Val} (5\%), and \textit{Test} (20\%) sets. 
Three additional splits challenge model generalization: \textit{Novel Object} (unseen objects in seen scenes), \textit{Novel Scene} (seen objects in unseen scenes), and \textit{Novel Scenario} (unseen objects in unseen scenes). 
Tables \ref{tab:benchmark_stats_1} and \ref{tab:benchmark_stats_2} present detailed statistics of these splits and task configurations, enabling systematic evaluation of model generalization across various dimensions of mobile manipulation in 3D scenes.

\textbf{Metrics.} \label{subsec:metrics}
We employ a multi-faceted approach to evaluate motion generation models. 
Task success rate serves as the primary metric, determined by the robot's ability to complete specified tasks and maintain the desired state for 2 seconds, as verified by the Isaac Sim physics engine. 
We also measure the closest distance from the end-effector to the target as an auxiliary metric, reflecting the trajectory's effectiveness in reaching the object or placement location. 
To assess trajectory quality, we utilize several quantitative measures: environment collision, self-collision, joint limit violation, and trajectory solving time.
This comprehensive set of metrics evaluates models' capabilities in generating effective and efficient motion trajectories for mobile manipulation in 3D scenes.

\section{Experiments}\label{sec:exp}

\subsection{Experimental Setup}

\textbf{Models for \benchmark.} Due to the lack of existing models for whole-body motion generation in mobile manipulation within 3D scenes, we adapt five state-of-the-art approaches to our benchmark:
\begin{itemize}
    \item \modmp: Integrates a \ac{vkc} motion planner~\cite{jiao2021consolidated} with grasp pose predictor~\cite{urain2023se} and heuristic placement.
    \item \modmps: Similar to \modmp, we replace the planer with a sampling-based planner RRT-Connect~\cite{kuffner2000rrt}.
    \item \mpnet~\cite{fishman2023motion}: Extended from stationary to mobile manipulation by incorporating whole-body joint generation and \ac{sdf}~\cite{wang2022dual} for collision loss computation in complex 3D scans.
    \item \mptf: A skill transformer~\cite{huang2023skill} variant using PointNet++~\cite{qi2017pointnet++} for 3D scan processing and decision transformer architecture~\cite{chen2021decision} for enhanced sequence modeling.
    \item \mdiffusion: Utilizes a conditional diffusion model~\cite{huang2023diffusion}, encoding 3D scans with a Point Transformer~\cite{zhao2021point} and employing a cross-attention module to predict denoising scores conditioned on 3D features.
\end{itemize}

\textbf{Implementation Details.} 
For \mpnet, \mdiffusion~and \mptf, we generate 3D scans from scene \ac{urdf}.
To enhance learning tractability, we apply a perception bounding box around the robot and target object to crop the scans, focusing the model's attention on relevant spatial information. 
We train \mpnet, \mdiffusion~and \mptf~on the \textit{Train} split and perform model selection on \textit{Val}. 
In constrast, as \modmp and \modmps~does not involve learning procedure, we evaluate it directly on the \textit{Test} and \textit{Novel} splits. 
To simplify the optimization problem in \modmp~and \modmps, we ignore collisions between the end-effector and target object during motion planning, as considering these collisions would frequently result in infeasible trajectories.

\subsection{Experimental Results}

\begin{table*}[ht!]
    \centering
    \caption{Quantitative results on \benchmark, measured by success rate (\textit{Succ}), distance to goal (\textit{Dist}), joint violation rate (\textit{J.Vio}), environment collision rate (\textit{E.Coll}), self-collision rate (\textit{S.Coll}), and execution time (\textit{Time}). Best performance is shown in bold.}
    \resizebox{\textwidth}{!}{%
    \setlength{\tabcolsep}{3pt}
    \setstretch{1.0}
    \begin{tabular}{cc >{\columncolor{red!20}}c >{\columncolor{red!20}}c >{\columncolor{red!20}}c >{\columncolor{red!20}}c >{\columncolor{red!20}}c >{\columncolor{red!20}}c >{\columncolor{blue!20}}c >{\columncolor{blue!20}}c >{\columncolor{blue!20}}c >{\columncolor{blue!20}}c >{\columncolor{blue!20}}c >{\columncolor{blue!20}}c}
    \toprule
    \multirow{2.5}{*}{\makecell[c]{Test Split}} & \multirow{2.5}{*}{\makecell[c]{Method}} & \multicolumn{6}{>{\columncolor{red!20}}c}{Pick Task} & \multicolumn{6}{>{\columncolor{blue!20}}c}{Place Task} \\
    \cmidrule(lr){3-8} \cmidrule(lr){9-14}
    & & Succ(\%)$\uparrow$ & Dist(m)$\downarrow$ & JVio(\%)$\downarrow$ & EnvColl(\%)$\downarrow$ & SelfColl(\%)$\downarrow$ & Time(s)$\downarrow$ & Succ(\%)$\uparrow$ & Dist(m)$\downarrow$ & JVio(\%)$\downarrow$ & EnvColl(\%)$\downarrow$ & SelfColl(\%)$\downarrow$ & Time(s)$\downarrow$ \\
    \midrule
    \multirow{5}{*}{\makecell[c]{\textit{Base}\\\textit{Test}}} 
    & \mpnet & $0.07$ & $0.34$ & $20.79$ & $16.53$ & $0.36$ & $\mathbf{0.48}$ & $0.80$ & $1.68$ & $34.67$ & $42.75$ & $1.24$ & $0.59$ \\
    & \mptf & $0.00$ & $1.36$ & $\mathbf{0.00}$ & $44.58$ & $\mathbf{0.00}$ & $0.93$ & $0.15$ & $0.92$ & $0.15$ & $23.38$ & $\mathbf{0.00}$ & $1.16$ \\
    & \mdiffusion & $18.12$ & $0.04$ & $0.59$ & $19.09$ & $0.53$ & $\mathbf{0.48}$ & $\mathbf{5.83}$ & $0.04$ & $0.36$ & $39.67$ & $0.32$ & $\mathbf{0.47}$ \\
    & \modmps & $16.90$ & $0.03$ & $\mathbf{0.00}$ & $12.13$ & $\mathbf{0.00}$ & $87.53$ & $1.98$ & $0.31$ & $\mathbf{0.00}$ & $3.58$ & $\mathbf{0.00}$ & $89.74$ \\
    & \modmp & $\mathbf{20.13}$ & $\mathbf{0.01}$ & $\mathbf{0.00}$ & $\mathbf{9.70}$ & $\mathbf{0.00}$ & $19.63$ & $2.76$ & $\mathbf{0.29}$ & $\mathbf{0.00}$ & $\mathbf{2.65}$ & $\mathbf{0.00}$ & $28.58$ \\
    \midrule
    \multirow{5}{*}{\makecell[c]{\textit{Novel}\\\textit{Object}}} 
    & \mpnet & $0.15$ & $0.34$ & $29.07$ & $22.38$ & $0.44$ & $0.47$ & $0.76$ & $1.55$ & $35.26$ & $45.84$ & $\mathbf{0.00}$ & $0.59$ \\
    & \mptf & $0.44$ & $1.39$ & $\mathbf{0.00}$ & $53.49$ & $\mathbf{0.00}$ & $0.94$ & $0.25$ & $0.70$ & $\mathbf{0.00}$ & $31.74$ & $\mathbf{0.00}$ & $1.16$ \\
    & \mdiffusion & $9.30$ & $0.05$ & $0.15$ & $37.24$ & $\mathbf{0.00}$ & $\mathbf{0.45}$ & $1.26$ & $0.06$ & $0.50$ & $35.32$ & $\mathbf{0.00}$ & $\mathbf{0.44}$ \\
    & \modmps & $18.39$ & $0.01$ & $\mathbf{0.00}$ & $19.75$ & $\mathbf{0.00}$ & $88.65$ & $3.41$ & $0.14$ & $\mathbf{0.00}$ & $4.53$ & $\mathbf{0.00}$ & $88.31$ \\
    & \modmp & $\mathbf{21.80}$ & $\mathbf{0.00}$ & $\mathbf{0.00}$ & $\mathbf{13.15}$ & $\mathbf{0.00}$ & $18.74$ & $\mathbf{5.10}$ & $\mathbf{0.12}$ & $\mathbf{0.00}$ & $\mathbf{0.00}$ & $\mathbf{0.00}$ & $29.89$ \\
    \midrule
    \multirow{5}{*}{\makecell[c]{\textit{Novel}\\\textit{Scene}}} 
    & \mpnet & $0.00$ & $0.42$ & $13.73$ & $43.88$ & $0.13$ & $\mathbf{0.48}$ & $0.84$ & $2.31$ & $41.78$ & $45.96$ & $4.18$ & $0.59$ \\
    & \mptf & $0.00$ & $2.06$ & $\mathbf{0.00}$ & $60.13$ & $\mathbf{0.00}$ & $0.93$ & $0.00$ & $1.04$ & $\mathbf{0.00}$ & $13.65$ & $\mathbf{0.00}$ & $1.17$ \\
    & \mdiffusion & $7.25$ & $0.04$ & $0.13$ & $38.53$ & $0.13$ & $\mathbf{0.48}$ & $1.95$ & $0.07$ & $0.28$ & $45.13$ & $\mathbf{0.00}$ & $\textbf{0.44}$ \\
    & \modmps & $19.20$ & $0.02$ & $\mathbf{0.00}$ & $13.27$ & $\mathbf{0.00}$ & $89.10$ & $7.80$ & $0.23$ & $\mathbf{0.00}$ & $3.49$ & $\mathbf{0.00}$ & $88.93$ \\
    & \modmp & $\mathbf{25.59}$ & $\mathbf{0.00}$ & $\mathbf{0.00}$ & $\mathbf{10.82}$ & $\mathbf{0.00}$ & $20.13$ & $\mathbf{9.76}$ & $\mathbf{0.18}$ & $\mathbf{0.00}$ & $\mathbf{1.10}$ & $\mathbf{0.00}$ & $27.39$ \\
    \midrule
    \multirow{5}{*}{\makecell[c]{\textit{Novel}\\\textit{Scenario}}} 
    & \mpnet & $0.00$ & $0.61$ & $16.67$ & $25.49$ & $\mathbf{0.00}$ & $0.47$ & $0.00$ & $2.74$ & $16.88$ & $9.09$ & $1.30$ & $0.59$ \\
    & \mptf & $0.00$ & $2.58$ & $\mathbf{0.00}$ & $70.59$ & $\mathbf{0.00}$ & $0.92$ & $0.00$ & $1.68$ & $9.09$ & $12.99$ & $\mathbf{0.00}$ & $1.17$ \\
    & \mdiffusion & $5.88$ & $0.04$ & $\mathbf{0.00}$ & $26.76$ & $\mathbf{0.00}$ & $\mathbf{0.46}$ & $2.60$ & $0.04$ & $\mathbf{0.00}$ & $7.49$ & $\mathbf{0.00}$ & $\mathbf{0.45}$ \\
    & \modmps & $20.12$ & $0.02$ & $\mathbf{0.00}$ & $14.97$ & $\mathbf{0.00}$ & $89.32$ & $4.31$ & $0.38$ & $\mathbf{0.00}$ & $2.67$ & $\mathbf{0.00}$ & $89.19$ \\
    & \modmp & $\mathbf{23.94}$ & $\mathbf{0.00}$ & $\mathbf{0.00}$ & $\mathbf{11.81}$ & $\mathbf{0.00}$ & $19.49$ & $\mathbf{6.52}$ & $\mathbf{0.25}$ & $\mathbf{0.00}$ & $\mathbf{0.00}$ & $\mathbf{0.00}$ & $28.31$ \\
    \bottomrule
    \end{tabular}%
    }%
    \label{tlb:exp}
\end{table*}

The experimental results are summarized in~\cref{tlb:exp}.
Trajectories are evaluated in Isaac Sim using metrics described in~\cref{subsec:metrics}.
Particularly, for the \modmp~and \modmps~model, when motion planning fails to solve the problem (\ie, optimization does not converge), we consider it as a failure instance.

\textbf{Across Models.} \modmp~consistently outperforms other approaches in most pick-and-place tasks, achieving higher success rates and closer goal distances. This confirms our hypothesis that integrating conventional motion planning with affordance prediction generalizes effectively across diverse 3D scenes. 
While the sampling-based \modmps~performs comparably to the optimization-based \modmp, it requires substantially more processing time. 
However, \modmp's superior performance comes with increased computation time due to optimization complexity in large-scale environments. 
Its effectiveness depends heavily on grasp and placement pose prediction quality; inaccuracies can lead to optimization failures or collisions (see~\cref{fig:teaser_fail_traj}). 
Despite promising results, the overall low success rates indicate that this integration alone is insufficient for robust performance.

In contrast, learning-based models are more time-efficient but often fail to generate feasible solutions in unseen scenarios. 
Specifically, while \mdiffusion~achieves  comparable performance to planning-based methods, its success drops significantly in the \textit{Novel} splits. 
Unlike planning-based methods with hard constraints, learning-based approaches frequently produce trajectories violating joint limitations and prone to collisions. 
\mpnet~and \mptf~achieve minimal success in test splits, suggesting that direct adaptation of stationary manipulation models to mobile manipulation is infeasible without sufficient model capacity to capture environmental complexity. 
These findings underscore the challenge of generating whole-body motion trajectories in complex 3D environments and highlight the need for more sophisticated mobile manipulation models.

\textbf{Across Tasks.} The experiment results reveal distinct performance patterns between pick and place tasks. 
While \modmp~maintains better performance in both tasks, its success rates significantly drop in place tasks, and all models require more time to generate trajectories for the \textit{place tasks}.
This discrepancy suggests that generating coordinated whole-body motion trajectories for placing objects is more challenging than for picking objects, as it involves additional constraints such as stable placement locations, appropriate object orientation, and reachable motion trajectory.
The increased complexity of place tasks explains the lower success rates and longer execution times observed across all models.

\textbf{On Generalization.} 
While \mdiffusion~achieves  comparable performance to planning-based methods in the \textit{Base} split, its performance deteriorates in the \textit{Novel} splits,indicating persistent generalization challenges for learning-based models.
Particularly, the distance to goal in unfamiliar scenes (\textit{Novel Scene} and \textit{Novel Scenario} splits) exceeds that of the \textit{Novel Object} split, suggesting that the impact of novel scenes is more significant than novel objects.
The conventional planning-based method, in contrast, maintains consistent performance across all splits, though its relatively low success rates across the board underscore the inherent complexity of mobile manipulation tasks in diverse household environments.


\textbf{Remarks.} Our experiments reveal two key insights:
\begin{itemize}
    \item Although combining motion planning with affordance prediction demonstrates consistent performance across all splits, its overall success rates remain low. This highlights the limitations of hybrid approaches and the need for holistic solutions to mobile manipulation in 3D scenes.
    \item Mobile manipulation requires models with greater expressiveness and generalization than stationary tasks. Poor performance in unseen scenarios (\ie, \textit{Novel} splits) underscores the need for (a) fine-grained perceptual representations and (b) advanced whole-body motion models for feasible trajectories.
\end{itemize}

\section{Conclusion}\label{sec:conclude}

We introduced \benchmark, a comprehensive benchmark for whole-body motion generation in mobile manipulation tasks across diverse 3D environments, featuring 30k object rearrangement tasks in 119 household scenes. 
\benchmark~provides a standardized platform for both planning and learning communities.
Through comprehensive evaluations of state-of-the-art models, we highlighted the persistent challenges in generating coordinated base-arm motion trajectories that satisfy both environmental constraints and task objectives. 
Furthermore, we developed \datatool, a tool designed to efficiently generate whole-body motion trajectories from high-level instructions, which can serve as a valuable resource for researchers in their own studies. 
We hope \benchmark~opens new opportunities for robotics research and catalyzes progress toward developing more adaptive and capable embodied agents.

\textbf{Acknowledgement:}
We thank the all the colleagues in Robotics Lab from BIGAI for fruitful discussions.

\balance
\setstretch{0.95}
\bibliographystyle{ieeetr}
\bibliography{reference_header,reference}

\end{document}